\documentclass[nohyperref]{article}

\usepackage{amsmath,amsfonts,bm}

\def\eqref#1{equation~\ref{#1}}
\def\1{\bm{1}}

\DeclareMathAlphabet{\mathsfit}{\encodingdefault}{\sfdefault}{m}{sl}
\SetMathAlphabet{\mathsfit}{bold}{\encodingdefault}{\sfdefault}{bx}{n}

\usepackage{microtype}
\usepackage{graphicx}
\usepackage{subcaption}
\usepackage[utf8]{inputenc} % allow utf-8 input
\usepackage[T1]{fontenc}    % use 8-bit T1 fonts
\usepackage{booktabs}
\usepackage{dblfloatfix} 
\usepackage{enumitem, kantlipsum}
\usepackage{tcolorbox}
\usepackage{siunitx}
\usepackage{xcolor}
\usepackage{nicefrac} 
\usepackage{multirow}
\usepackage{threeparttable}
\usepackage[framemethod=tikz]{mdframed}

\usepackage{hyperref}

\usepackage[accepted]{icml2022}

\usepackage{amsmath}
\usepackage{amssymb}
\usepackage{mathtools}
\usepackage{amsthm}

\usepackage[capitalize,noabbrev]{cleveref}

\theoremstyle{plain}
\newtheorem{theorem}{Theorem}[section]

\theoremstyle{definition}
\newtheorem{definition}[theorem]{Definition}

\theoremstyle{remark}

\newcommand*\Diff{\mathop{}\!\mathrm{d}}
\newcommand{\dtrain}{\ensuremath{\mathrm{train}}}
\newcommand{\dval}{\ensuremath{\mathrm{val}}}
\newcommand{\dtest}{\ensuremath{\mathrm{test}}}
\newcommand{\CE}{\ensuremath{\mathrm{CE}}}
\newcommand{\util}{\ensuremath{\mathrm{util}}}
\newcommand{\speed}{\ensuremath{\mathrm{speed}}}

\newcommand{\depth}{\ensuremath{\mathrm{depth}}}
\newcommand{\rgb}{\ensuremath{\mathrm{RGB}}}

\definecolor{mycolor}{rgb}{0.122, 0.435, 0.698}
\newmdenv[innerlinewidth=0.8pt, roundcorner=2pt,linecolor=mycolor,innerleftmargin=3pt,
innerrightmargin=3pt,innertopmargin=3pt,innerbottommargin=3pt]{mybox}
\newtcolorbox{mybox_}{colback=gray!5!white,colframe=gray!75!black}

\usepackage[textsize=tiny]{todonotes}

\icmltitlerunning{Characterizing and overcoming the greedy nature of learning in multi-modal deep neural networks}

\begin{document}

\twocolumn[
\icmltitle{Characterizing and Overcoming the Greedy Nature of Learning in Multi-modal Deep Neural Networks}

% It is OKAY to include author information, even for blind
% submissions: the style file will automatically remove it for you
% unless you've provided the [accepted] option to the icml2022
% package.

% List of affiliations: The first argument should be a (short)
% identifier you will use later to specify author affiliations
% Academic affiliations should list Department, University, City, Region, Country
% Industry affiliations should list Company, City, Region, Country

% You can specify symbols, otherwise they are numbered in order.
% Ideally, you should not use this facility. Affiliations will be numbered
% in order of appearance and this is the preferred way.
\icmlsetsymbol{equal}{*}

\begin{icmlauthorlist}
\icmlauthor{Nan Wu}{1}
\icmlauthor{Stanisław Jastrzębski}{2,1}
\icmlauthor{Kyunghyun Cho}{1,3,4,5}
\icmlauthor{Krzysztof J. Geras}{2,1,3}
\end{icmlauthorlist}

\icmlaffiliation{1}{Center for Data Science, New York University}
\icmlaffiliation{2}{NYU Grossman School of Medicine}
\icmlaffiliation{3}{Courant Institute of Mathematical Sciences, New York University}
\icmlaffiliation{4}{Genentech}
\icmlaffiliation{5}{CIFAR LMB}

\icmlcorrespondingauthor{Nan Wu}{nan.wu@nyu.edu}

% You may provide any keywords that you
% find helpful for describing your paper; these are used to populate
% the "keywords" metadata in the PDF but will not be shown in the document
\icmlkeywords{Machine Learning, ICML}

\vskip 0.3in
]

% this must go after the closing bracket ] following \twocolumn[ ...

% This command actually creates the footnote in the first column
% listing the affiliations and the copyright notice.
% The command takes one argument, which is text to display at the start of the footnote.
% The \icmlEqualContribution command is standard text for equal contribution.
% Remove it (just {}) if you do not need this facility.

%\printAffiliationsAndNotice{}  % leave blank if no need to mention equal contribution
\printAffiliationsAndNotice{} % otherwise use the standard text.
\begin{abstract}
We hypothesize that due to the greedy nature of learning in multi-modal deep neural networks, these models tend to rely on just one modality while under-fitting the other modalities. Such behavior is counter-intuitive and hurts the models' generalization, as we observe empirically. To estimate the model's dependence on each modality, we compute the gain on the accuracy when the model has access to it in addition to another modality. We refer to this gain as the \emph{conditional utilization rate}. In the experiments, we consistently observe an imbalance in conditional utilization rates between modalities, across multiple tasks and architectures. Since conditional utilization rate cannot be computed efficiently during training, we introduce a proxy for it based on the pace at which the model learns from each modality, which we refer to as the \emph{conditional learning speed}. We propose an algorithm to balance the conditional learning speeds between modalities during training and demonstrate that it indeed addresses the issue of greedy learning.\footnote{We provide an implementation of the proposed methods at  \href{https://github.com/nyukat/greedy_multimodal_learning}{https://github.com/nyukat/greedy\_multimodal\_learning}} The proposed algorithm improves the model's generalization on three datasets: Colored MNIST, ModelNet40, and NVIDIA Dynamic Hand Gesture. 
\end{abstract}

\section{Introduction}
In real-world problems, each instance frequently has multiple modalities associated with it. For example, we detect cancer in both X-ray and ultrasound images. We seek clues from images to answer questions given in text. We are naturally interested in training deep neural networks (DNNs) end-to-end to learn from all available input modalities. We refer to such a training regime as a \emph{multi-modal learning process} and DNNs resulting from it as \emph{multi-modal DNNs}.

Several recent studies have reported unsatisfactory performance of multi-modal DNNs in various tasks~\citep{wang2020makes, wu2020improving, NEURIPS2020_20d749bc, cadene2019rubi, agrawal2016analyzing, hessel2020does, han2021trusted, hessel2020does}. For example, in Visual Question Answering (VQA), multi-modal DNNs were found to ignore the visual modality and exploit statistical regularities shared between the text in the question and the text in the answer alone, resulting in poor generalization~\citep{cadene2019rubi, NEURIPS2020_20d749bc, agrawal2016analyzing}. Similarly, in Human Action Recognition, multi-modal DNNs trained on images and audio were observed to perform worse than uni-modal DNNs trained on images only~\citep{wang2020makes}. In addition to the multi-modal classifiers, the unbalanced modality-wise utilization has been identified in the multi-modal pre-trained models~\citep{li2020closer, cao2020behind}.

These earlier negative findings compel us to ask, \emph{what prevents multi-modal DNNs from achieving better performance?} In order to answer this question, we put forward the \emph{greedy learner hypothesis}. The greedy learner hypothesis states that a multi-modal DNN learns to rely on one of the input modalities, based on which it could learn faster, and does not continue to learn to use the other modalities. This greediness prevents a multi-modal DNN from learning to exploit all available modalities and often results in worse generalization. It explains the challenge in training multi-modal DNNs and motivates us to design a better multi-modal learning algorithm.  %leads to an imbalance in conditional utilization rates between modalities. In other words, it

We first diagnose a multi-modal DNN's ability to utilize all modalities by analyzing its \emph{conditional utilization rates}. On a task with two modalities, $m_0$ and $m_1$, we define a model's conditional utilization rates as the relative difference in accuracy between two models derived from the DNN, one using both modalities and the other using only one modality. The conditional utilization rate of $m_1$ given $m_0$, denoted by $\bm{u}(m_1|m_0)$, measures how important it is for the model to use $m_1$ in order to reach accurate predictions, given the presence of $m_0$. In several multi-modal learning tasks, we consistently observe a significant imbalance in conditional utilization rate between modalities. For example, we have $\bm{u}(\depth|\rgb)=0.63$ and $\bm{u}(\rgb|\depth)=0.01$ for a model trained for Hand Gesture Recognition using the RGB and the depth modalities. Since $0\leq|\bm{u}|\leq1$, these two observed values indicate that the model solely relies on the depth modality, ignoring the RGB modality almost completely. These observations support the conjecture that the multi-modal learning process often results in models that under-utilize some of the input modalities.  

According our proposed greedy learner hypothesis, it is the different speeds at which a multi-modal DNN learns from different modalities that leads to an imbalance in conditional utilization rate. If we intervene in the training process to adjust these speeds, we may be able to prevent the hurtful imbalance across input modalities. We analyze the learning dynamics of model components and propose a metric called \emph{conditional learning speed}, defined using the gradient norm and the norm of models' parameters. It measures the speed at which the model learns from one modality relative to the other modalities. We empirically show that it is a reasonable proxy for conditional utilization rate. We introduce a training algorithm called \emph{balanced multi-modal learning} which uses conditional learning speed to guide the model to learn from previously underutilized modalities. We show that models trained with this algorithm learn to use all modalities appropriately and achieve stronger generalization on three multi-modal datasets: Colored MNIST dataset~\citep{kim2019learning}, ModelNet40 dataset of 3D objects~\citep{su15mvcnn} and NVGesture dataset~\citep{molchanov2015hand}.

\section{Related Work}
\paragraph{Dataset and model bias inspection in VQA}
Multi-modal DNNs are expected to leverage cross-modal interactions in VQA, but have been reported to exploit the modality-wise bias in the data~\citep{jabri2016revisiting, balanced_vqa_v2, winterbottom2020modality}. Several frameworks were proposed to evaluate the severity of this issue, such as constructing de-biased datasets~\citep{hessel2020does, agrawal2018don}, evaluating models' performance when eliminating cross-modal interactions via empirical function projection~\citep{hessel2020does} or via permuting the features of a modality across samples~\citep{gat2021perceptual}. There are also efforts to overcome this problem. Many of them rely on a question-only branch for capturing spurious relationships between questions and answer candidates and help through cross-entropy re-weighing strategies~\citep{cadene2019rubi, han2015greedy, lao2021superficial}. 
Instead of estimating the bias through a question-only branch, \citet{NEURIPS2020_20d749bc} proposes to supply inputs with Gaussian perturbations to the model and regularize it by maximizing functional entropies, in order to force the model to use multiple sources of information. In this work, we expand the discussion from VQA to multi-modal classification tasks. These tasks are different from VQA but the problem identified above also appears there. We explain this phenomenon by the greedy nature of learning in multi-modal DNNs and design tools to overcome inadequate modality utilization in multi-modal classification. 

\paragraph{Video classification}
How to benefit from multi-modality in video data has been a popular research direction for video understanding. Prior work focuses primarily on improving architectural designs of the DNNs~\citep{ngiam2011multi-modal, neverova2015moddrop, tran2015learning, lazaridou2015combining, perez2019mfas, joze2020mmtm, arnab2021vivit, sun2021learning}. Our study is related to a recent study by~\citet{wang2020makes}. They observe that the best uni-modal DNNs often outperforms the multi-modal DNNs, and propose a framework that estimates the uni-modal branches' generalization and overfitting speeds in order to calibrate the learning through loss re-weighing. Their proposed methods rely on estimating models' performance occasionally on a held-out validation set during training, which makes it costly and unstable for small datasets. To serve a similar purpose but not from the perspective of model optimization, \citet{neurips-wang2020deep} design a parameter-free multi-modal fusion framework that dynamically exchanges channels between the uni-modal branches. In this work, we provide tools to directly examine the imbalanced modality utilization in jointly-trained multi-modal DNNs, and implement calibration through a computationally efficient method. We focus on verifying our solution with intermediate fusion as it yields better performance than late fusion~\citep{joze2020mmtm}, though an investigation such as ours was missing in the literature.

\section{Problem Setup}
\label{sec:formu}
Without loss of generality, we consider two input modalities, referred to as $m_0$ and $m_1$. We denote a multi-modal dataset by $\mathcal{D}$. This dataset consists of multiple instances $(\bm{x}, y)$, where $\bm{x}=(\bm{x}_{m_0}, \bm{x}_{m_1})$. We partition the dataset $\mathcal{D}$ into training, validation and test sets, denoted by $\mathcal{D}^{\dtrain}$, $\mathcal{D}^{\dval}$, and $\mathcal{D}^{\dtest}$, respectively. The goal is to use this data set to train a model that accurately predicts $y$ from $\bm{x}$.

We use a multi-modal DNN with two uni-modal branches, denoted by $\phi_{0}$ and $\phi_{1}$, taking $\bm{x}_{m_0}$ and $\bm{x}_{m_1}$ as input. 
The two uni-modal branches are interconnected by layer-wise fusion modules. According to the categorization of fusion strategies in the deep learning literature, this is considered ``intermediate fusion''~\citep{ngiam2011multi-modal,atrey2010multimodal,baltruvsaitis2018multimodal}. It has demonstrated competitive performance in comparison to multi-modal DNNs with late fusion in multiple tasks~\citep{perez2018film, joze2020mmtm, anderson2018bottom,neurips-wang2020deep}.

Specifically, we implement each fusion module with a multi-modal transfer module (MMTM)~\citep{joze2020mmtm}. It connects corresponding convolutional layers from different uni-modal branches. Let $\bm{A}_{0} \in \mathbb{R}^{N_1\times \dots \times N_L \times C}$ and $\bm{A}_{1} \in \mathbb{R}^{M_1\times \dots \times M_J \times C'}$ denote feature maps produced by these two layers. Here, $N_i$ and $M_i$ represent the spatial dimensions of each feature map, and $C$ and $C'$ represent the number of feature maps. We apply global average pooling over spatial dimensions of the feature maps and get two vectors, denoted by $h_{0}\in \mathbb{R}^{C}$ and $h_{1}\in \mathbb{R}^{C'}$. A fusion mechanism within MMTM is then applied to the two vectors:
\begin{equation}
    w_{0}, w_{1} = g([h_{0},\, h_{1}])
\end{equation} where $[\cdot,\cdot]$ represents the concatenation operation; $w_0\in \mathbb{R}^{C}$ and $w_1\in \mathbb{R}^{C'}$. The function $g$ is parameterised with a fully-connected layer, ReLu~\citep{nair2010rectified} function, and two more fully-connected layers, each returning an activation in the dimension of $C$ and $C'$. 

Next, we re-scale the feature maps ($\bm{A}_{0}$ and $\bm{A}_{1}$) with the obtained activations ($w_0$ and $w_1$):
\begin{align}
    \Tilde{\bm{A}_0} &= 2 \times \sigma(w_0) \odot \bm{A}_0,  &\Tilde{\bm{A}_1} &= 2 \times \sigma(w_1) \odot \bm{A}_1
\end{align}
where $\odot$ is the channel-wise product operation and $\sigma(\cdot)$ is the sigmoid function. 

We feed $\Tilde{\bm{A}_{0}}$ and $\Tilde{\bm{A}_{1}}$ to the next layer of $\phi_0$ and $\phi_1$. Thus information from one modality is shared from one uni-modal branch to the other.  

\begin{figure}[t!]
\centering
\includegraphics[width=0.99\linewidth, trim ={0, 0, 0, 2mm}, clip]{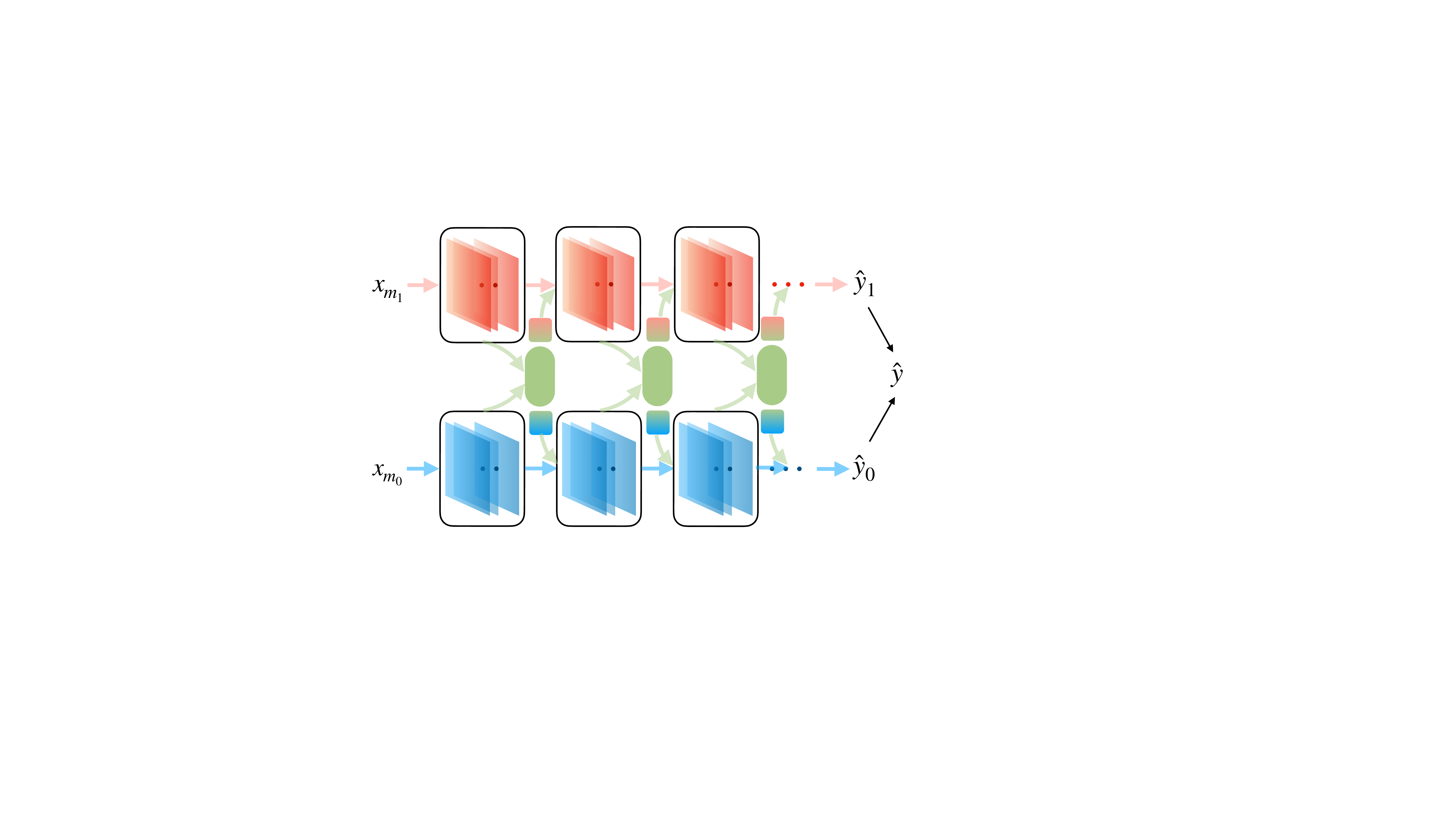}
\caption{The multi-modal DNN with intermediate fusion that we study in this work. Blue and red blocks represent the two uni-modal networks, $\phi_0$ and $\phi_1$, learning from modalities $m_0$ and $m_1$. The fusion layers are marked with green, green-blue, and green-red. We denote the predictions from $\phi_0$ and $\phi_1$ by $\hat{y}_{0}$ and $\hat{y}_{1}$, and the average of the two by $\hat{y}$ which is the model's prediction.}\label{fig:info-flow-multi-modal}
\end{figure}

We train a multi-modal DNN $f$ on $\mathcal{D}^{\dtrain}$. Let $\hat{y}$ be the prediction of $f$ for an input $\bm{x} =(\bm{x}_{m_0}, \bm{x}_{m_1})$: \begin{equation}\hat{y}= f(\bm{x}_{m_0}, \bm{x}_{m_1}).\end{equation} As shown in Figure~\ref{fig:info-flow-multi-modal}, $\hat{y}= \frac{1}{2} (\hat{y}_{0} + \hat{y}_{1})$, where $\hat{y}_{0}$ and $\hat{y}_{1}$ are the outputs of the two uni-modal branches. 

During training, the parameters of $f$ are updated by stochastic gradient descent (SGD) to minimize the loss: \begin{equation}L=\CE(y, \hat{y}_{0}) + \CE(y, \hat{y}_{1})\end{equation} where $\CE$ stands for cross-entropy. We refer to each of the cross-entropy losses as a modality-specific loss. We train the model until it reaches 100\% classification accuracy on $\mathcal{D}^{\dtrain}$, and pick the model checkpoint associated with the highest validation accuracy across all epochs.

\section{The Greedy Learner Hypothesis}

In this section, we first derive conditional utilization rate to help explain the multi-modal DNN and to quantify its imbalance in modality utilization. Then we introduce the greedy learner hypothesis to explain challenges observed in training multi-modal DNNs. 

\subsection{Conditional Utilization Rate}
In the multi-modal DNN shown in \cref{fig:info-flow-multi-modal}, each uni-modal branch largely focuses on the associated input modality, and the fusion modules generate context information using all modalities, and feed the information to the uni-modal branches. Both $\hat{y}_{0}$ and $\hat{y}_{1}$ depend on information from both modalities. We derive $f_{0}$ and $f_{1}$ from $f$:
\begin{equation}\hat{y}_{0}=f_{0}(\bm{x}_{m_0}, \bm{x}_{m_1}),\quad \hat{y}_{1}=f_{1}(\bm{x}_{m_0}, \bm{x}_{m_1}).\end{equation}

Let $\mathcal{D}^{\dtrain} = \{\bm{x}^i\}_{i=1}^{n}$ and $h_0(\bm{x}^i)$ and $h_1(\bm{x}^i)$ denote the value of $h_0$ and $h_1$ when the model takes $\bm{x}^i$ as input. We compute the average of $h_0$ and $h_1$ over $\mathcal{D}^{\dtrain}$: \begin{equation}\overline{h}_0 = \frac{1}{n}\sum_{i=1}^{n}{h_0(\bm{x}^i)}, \quad \overline{h}_1 = \frac{1}{n}\sum_{i=1}^{n}{h_1(\bm{x}^i)}.\end{equation}

To estimate $f$'s reliance on each modality independently, we modify the operations introduced in \S\ref{sec:formu} for each fusion module as below: 
\begin{equation}w_0, * = g([h_0,\, \overline{h}_1]), \quad *, w_1 = g([\overline{h}_0,\, h_1]).\end{equation}

With such modified operation in every fusion module in the multi-modal DNN, we cut off the road to share information between the uni-modal branches and let the output of each branch rely on a single modality. We derive $f'_{0}$ and $f'_{1}$ and denote their outputs by $\hat{y}'_{0}$ and $\hat{y}'_{1}$:
\begin{equation}\hat{y}'_{0}=f'_{0}(\bm{x}_{m_0}),\quad \hat{y}'_{1}=f'_{1}(\bm{x}_{m_1}).\end{equation}

In summary, we derive four models from $f$ and compute the accuracy of each on $\mathcal{D}^{\dtest}$, denoted by $A(\cdot)$. We group the four accuracies into two pairs: $(A(f_0), A(f'_0))$, and $(A(f_1), A(f'_1))$. 

We define the conditional utilization rates as bellow.
\begin{definition}[Conditional utilization rate] 
For a multi-modal DNN, $f$, taking two modalities $m_0$ and $m_1$ as inputs, its conditional utilization rates for $m_0$ and $m_1$ are defined as \begin{equation}
\bm{u}(m_0|m_1)=\frac{A(f_1)-A(f'_1)}{A(f_1)},
\end{equation}
and
\begin{equation}\bm{u}(m_1|m_0)= \frac{A(f_0)-A(f'_0)}{A(f_0)}.
\end{equation}
\end{definition}
The conditional utilization rate is the relative change in accuracy between the two models within each pair. For example, $u(m_0|m_1)$ measures the marginal contribution that 
$m_0$ has in increasing the accuracy of $\hat{y}_{1}$. 

Let $\Diff_{\util}(f)$ denote the difference between conditional utilization rates: $\Diff_{\util}(f) = \bm{u}(m_1|m_0) - \bm{u}(m_0|m_1).$ It is bounded between $-1$ and $1$. When $\Diff_{\util}(f)$ is close to $-1$ or $1$, the model benefits only from one of the modalities given the other but not vice versa. This implies that the model's ability to predict $\hat{y}_{0}$ and $\hat{y}_{1}$ comes only from one of the modalities. Thus, if we observe high $|\Diff_{\util}(f)|$, we say that $f$ exhibits imbalance in utilization between modalities. 

\subsection{Multi-modal Learning Process is Greedy}
\label{sec:greedy}
We propose our hypothesis that builds on the following assumptions regarding multi-modal data, as well as observations previously made in the literature.

First, for most multi-modal learning tasks, both modalities are assumed to be predictive of the target. This can be expressed as $I(\bm{Y}, \bm{X}_{m_0})>0$ and $I(\bm{Y}, \bm{X}_{m_1})>0$, where $I$ denotes mutual information~\citep{blum1998combining, sridharan2008information}. In order to minimize one of the modality-specific losses, e.g. $\CE(y, \hat{y}_0)$, one can either update the parameters of the uni-modal branch taking $m_0$ as input, or the parameters of the fusion layers that pass information from $m_1$ to $\hat{y}_{0}$, or both.

Second, as shown in many tasks, different modalities have been said to be predictive of the target at different degrees. For example, it has been observed that when training DNNs on each modality separately, they usually do not reach the same level of performance~\citep{neurips-wang2020deep, joze2020mmtm}. In addition, multi-modal DNNs exhibit varying accuracy when trained on different subsets of modalities present for the task~\citep{weng2021hms, perez2019mfas, liu2018learn}.         
Third, DNNs learn from different modalities at different speeds. This has been observed in both uni-modal DNNs and multi-modal DNNs~\citep{wang2020makes, wu2020improving}.

Now we formulate the \emph{greedy learner hypothesis} to explain the behavior of multi-modal DNNs:
\begin{mybox_}
A multi-modal learning process is \emph{greedy} when it produces models that rely on only one of the available modalities. The modality that the multi-modal DNN primarily relies on is the modality that is the fastest to learn from. We {\it hypothesize} that a multi-modal learning process, in which a multi-modal DNN is trained to minimize the sum of modality-specific losses, is greedy.  
\end{mybox_}

To characterize greediness of the multi-modal learning process, we need to repeat it several times. Every time, we sample a learning rate from a given range and initialize the network' parameters randomly. Let $\mathbb{E}[\widehat{\Diff}_{\util}]$ denote the expectation of $\widehat{\Diff}_{\util}$ over the empirical distribution of the models obtained. The absolute value of $\mathbb{E}[\widehat{\Diff}_{\util}]$ is associated with the greediness of the process. The higher the $|\mathbb{E}[\widehat{\Diff}_{\util}]|$, the greedier the learning process.

To validate our hypothesis empirically, we conduct experiments on several multi-modal datasets using different network architectures (cf. \S\ref{sec:exp}). We show that the multi-modal learning process cannot avoid producing models that exhibit a high degree of imbalance in utilization between modalities. Their performance can be improved by using a less greedy algorithm which we will introduce in the next section.

\section{Making Multi-modal Learning Less Greedy}
We aim to make multi-modal learning less greedy by controlling the speed at which a multi-modal DNN learns to rely on each modality. In order to achieve this, we first define conditional learning speed to measure the speed at which the DNN learns from one modality. It serves as an efficient proxy for the conditional utilization rate of the corresponding modality, as shown empirically in \S\ref{sec:diff} and \S\ref{sec:reg}. We then propose the balanced multi-modal learning algorithm, which controls the difference in conditional learning speed between modalities that the model exhibits during training.

\subsection{Conditional Learning Speed}
As demonstrated in \S\ref{sec:greedy}, the imbalance in conditional utilization rates is a sign of the model exploiting the connection between the target and only one of the input modalities, ignoring cross-modal information. However, conditional utilization rates are designed to be measured after training is done and they are derived from the model's accuracy achieved on the held-out test set. This makes conditional utilization rate not the best tool to use to monitor training at the real-time. We instead derive a proxy metric, called \emph{conditional learning speed}, that captures relative learning speed between modalities during training.

Let us revisit the multi-modal DNN presented in Figure~\ref{fig:info-flow-multi-modal}. We denote the parameters of uni-modal branches $\phi_{0}$ and $\phi_{1}$ by $\bm{\theta}_{0}$ and $\bm{\theta}_{1}$. Besides $\bm{\theta}_{0}$, we consider the parameters of the layers marked with green and green/blue, as important components of the function mapping $\bm{x}$ to $\hat{y}_0$ (i.e., $f_0$). These parameters are from the fusion modules and we refer to them as $\bm{\theta}'_{0}$. Analogously, parameters in the layers marked with green and green/red are considered as important contributors to the function mapping $\bm{x}$ to $\hat{y}_1$ (i.e., $f_1$). We denote them by $\bm{\theta}'_{1}$. In this way, we divide the model's parameters into two pairs: $(\bm{\theta}_{0}, \bm{\theta}'_{0})$ and $(\bm{\theta}_{1}, \bm{\theta}'_{1})$.\footnote{
Note that parameters in the green block are both in $\bm{\theta}'_{0}$ and $\bm{\theta}'_{1}$.
}

We define the model's conditional learning speed as follows.
\begin{definition}[Conditional learning speed] 
Given a multi-modal DNN, $f$, with two input modalities, $m_0$ and $m_1$, the conditional learning speeds of these modalities after $t$ training steps, are
\begin{equation}
\bm{s}(m_1|m_0; t) = \log \frac{\sum_{i=1}^{t} \mu(\bm{\theta}'_0; i)}{\sum_{i=1}^{t} \mu(\bm{\theta}_0; i)}, 
\end{equation} 
and
\begin{equation}
\bm{s}(m_0|m_1; t) = \log \frac{\sum_{i=1}^{t} \mu(\bm{\theta}'_1; i)}{\sum_{i=1}^{t} \mu(\bm{\theta}_1; i)},
\end{equation} 
where for any parameters $\bm{\theta}$, let $\bm{\theta}_{(i-1)}$ and $\bm{\theta}_{(i)}$ denote its value before and after the gradient descent step $i$; we have $\bm{G} = \frac{\partial{L}}{\partial{\bm{\theta}}}|_{\bm{\theta}_{(i-1)}}$; and $\mu(\bm{\theta}; i)=||\bm{G}||^2_2 / ||\bm{\theta}_{(i)}||^2_2$ quantifies the change of $f$ that comes from updating $\bm{\theta}$ at the $i^{th}$ step, which is also interpreted as the effective update on $\bm{\theta}$.  
\end{definition}

This definition of $\mu(\bm{\theta}; i)$ is inspired by the discussion on the effective update of parameters~\citep{van2017l2, zhang2018three, brock2021high, hoffer2018norm}. When normalization techniques, such as batch normalization~\citep{ioffe2015batch}, are applied to the DNNs, the key property of the weight vector, $\bm{\theta}$, is its direction, i.e.,  $\bm{\theta} / ||\bm{\theta}||^2_2$. Thus, we measure the update on $\bm{\theta}$ using the norm of its gradient normalized by its norm. 

The conditional learning speed, $\bm{s}(m_1|m_0; t)$, is the log-ratio between the learning speed of $\bm{\theta}'_{0}$ and that of $\bm{\theta}_{0}$. Because $\bm{\theta}'_{0}$ carries information from $m_1$ to $\hat{y}_{0}$ and information from $m_0$ to $\hat{y}_{0}$ is carried by $\bm{\theta}_{0}$, $\bm{s}(m_1|m_0; t)$ reflects how fast the model learns from $m_1$ relative to $m_0$ after the first $t$ steps.  

We compute the difference between $\bm{s}(m_1|m_0; t)$ and $\bm{s}(m_0|m_1; t)$ as: $
\Diff_{\speed}(f; t) = \bm{s}(m_1|m_0; t)-\bm{s}(m_0|m_1; t)$
analogous to $\Diff_{\util}(f)$. For each model, we report $\Diff_{\speed}(f; T)$ as $\Diff_{\speed}(f)$ where the model takes $T$ steps until reaching the highest accuracy on $\mathcal{D}^{\dval}$.

We anticipate that the conditional learning speed serves as a proxy for the conditional utilization rate. In \S\ref{sec:diff} and \S\ref{sec:reg}, we show the distributions of $\widehat{\Diff}_{\speed}$ and $\widehat{\Diff}_{\util}$ side-by-side to validate this. %We assume that, $\Diff_{\speed}(f; t)$, which we can observe during training, is predictive of $\Diff_{\util}(f)$ we observe at the end of training. 

\subsection{Balanced Multi-modal Learning}
\label{sec:guided_algorithm}

Since we assume $\Diff_{\speed}(f, t)$ is predictive of the imbalanced utilization between modalities, we can take advantage of $\Diff_{\speed}(f, t)$ to balance conditional utilization on-the-fly. In addition to training the network normally with both modalities, we accelerate the model to learn from either modality alternately to balance their conditional learning speeds. See \cref{alg:mitigate} for an overall description. 

\begin{algorithm}[!htb]
   \caption{Balanced Multi-modal Learning}
   \label{alg:mitigate}
\begin{algorithmic}
   \STATE {\bfseries Input:} re-balancing window size $Q$, imbalance tolerance parameter $\alpha$,  \# epochs $N$, \# updating steps per epoch $n$
    \STATE {\bfseries Initiate:} $M_{\bm{\theta}_0}, M_{\bm{\theta}'_0}, M_{\bm{\theta}_1}, M_{\bm{\theta}'_1} \gets 0,0,0,0$; $q \gets Q$
   \FOR{$i=1$ {\bfseries to} $n$} 
    \STATE Take a regular step;
    \STATE $M_{\bm{\theta}_0}\gets M_{\bm{\theta}_0} + \mu(\bm{\theta}_0; i)$; $M_{\bm{\theta}'_0}\gets M_{\bm{\theta}'_0} + \mu(\bm{\theta}'_0; i)$; 
    \STATE $M_{\bm{\theta}_1}\gets M_{\bm{\theta}_1} + \mu(\bm{\theta}_1; i)$; $M_{\bm{\theta}'_1}\gets M_{\bm{\theta}_1} + \mu(\bm{\theta}'_1; i)$;
    \ENDFOR
    \FOR{$i=1$ {\bfseries to} $n\times N$}
    \IF{\textit{$q==Q$}} 
            \STATE Take a regular step;
            \STATE $M_{\bm{\theta}_0}\gets M_{\bm{\theta}_0} + \mu(\bm{\theta}_0; i)$; $M_{\bm{\theta}'_0}\gets M_{\bm{\theta}'_0} + \mu(\bm{\theta}'_0; i)$; 
    \STATE $M_{\bm{\theta}_1}\gets M_{\bm{\theta}_1} + \mu(\bm{\theta}_1; i)$; $M_{\bm{\theta}'_1}\gets M_{\bm{\theta}_1} + \mu(\bm{\theta}'_1; i)$;
            \STATE $\Diff_{\speed} = \log (M_{\bm{\theta}'_0}/M_{\bm{\theta}_0}) - \log (M_{\bm{\theta}'_1}/M_{\bm{\theta}_1})$;
            \STATE \textbf{if} $|\Diff_{\speed}|>\alpha$ \textbf{then} $q\gets 1$;
   \ELSE 
   \STATE $q\gets q+1$
        \STATE Take a re-balancing step to accelerate learning from $m_0$ \textbf{if} $\Diff_{\speed}>0$ \textbf{else} from $m_1$;
    \ENDIF
    \ENDFOR
\end{algorithmic}
\end{algorithm}

We refer to the training steps at which we perform forward and backward passes normally as the \emph{regular steps}. We introduce the \emph{re-balancing steps} at which we update one of the uni-modal branches intentionally in order to accelerate the model to learn from its input modality. 

In a re-balancing step where we aim to accelerate the model's learning from $m_0$, in every fusion module in the network, we re-scale the feature maps $\bm{A}_1$ in the same way as in \S\ref{sec:formu} but we re-scale $\bm{A}_0$ differently: 
\begin{align}
    \Tilde{\bm{A}_0} &= 2 \times \sigma(\overline{w}_0) \odot \bm{A}_0,  &
    \Tilde{\bm{A}_1} &= 2 \times \sigma(w_1) \odot \bm{A}_1,
\end{align}
where $\{\bm{x}^i\}_{i=1}^{n_t}$ are the $n_t$ samples used in the previous regular training steps; $w_0(\bm{x}^i)$ denotes the activation when the model takes $\bm{x}^i$ as input; and $\overline{w}_0 = 1/n_t \sum_{i=1}^{n_t}w_0(\bm{x}^i)$. 

In a re-balancing step to accelerate learning from $m_1$, we apply the modification on $\bm{A}_1$ instead of $\bm{A}_0$:
\begin{align}
    \Tilde{\bm{A}_0} &= 2 \times \sigma(w_0) \odot \bm{A}_0,  &
    \Tilde{\bm{A}_1} &= 2 \times \sigma(\overline{w}_1) \odot \bm{A}_1,
\end{align}
where $\overline{w}_1 = 1/n_t \sum_{i=1}^{n_t}w_1(\bm{x}^i).$

To warm-up the model, we perform only regular steps in the first epoch. Then we switch from regular steps to re-balancing steps if $|\Diff_{\speed}(t)|>\alpha$, where $\alpha$ is a hyperparameter, referred to as the \emph{imbalance tolerance parameter}. Once it switches to re-balancing mode, we takes $Q$ re-balancing steps before returning to the regular mode. We refer to the hyperparameter $Q$ by the \emph{re-balancing window size}.   

\begin{figure*}[h]
 \vspace*{\fill}
  \centering
       \begin{subfigure}[b]{0.48\textwidth}
         \centering
         \includegraphics[width=1\textwidth, trim={10mm 10mm 5mm 5mm},clip]{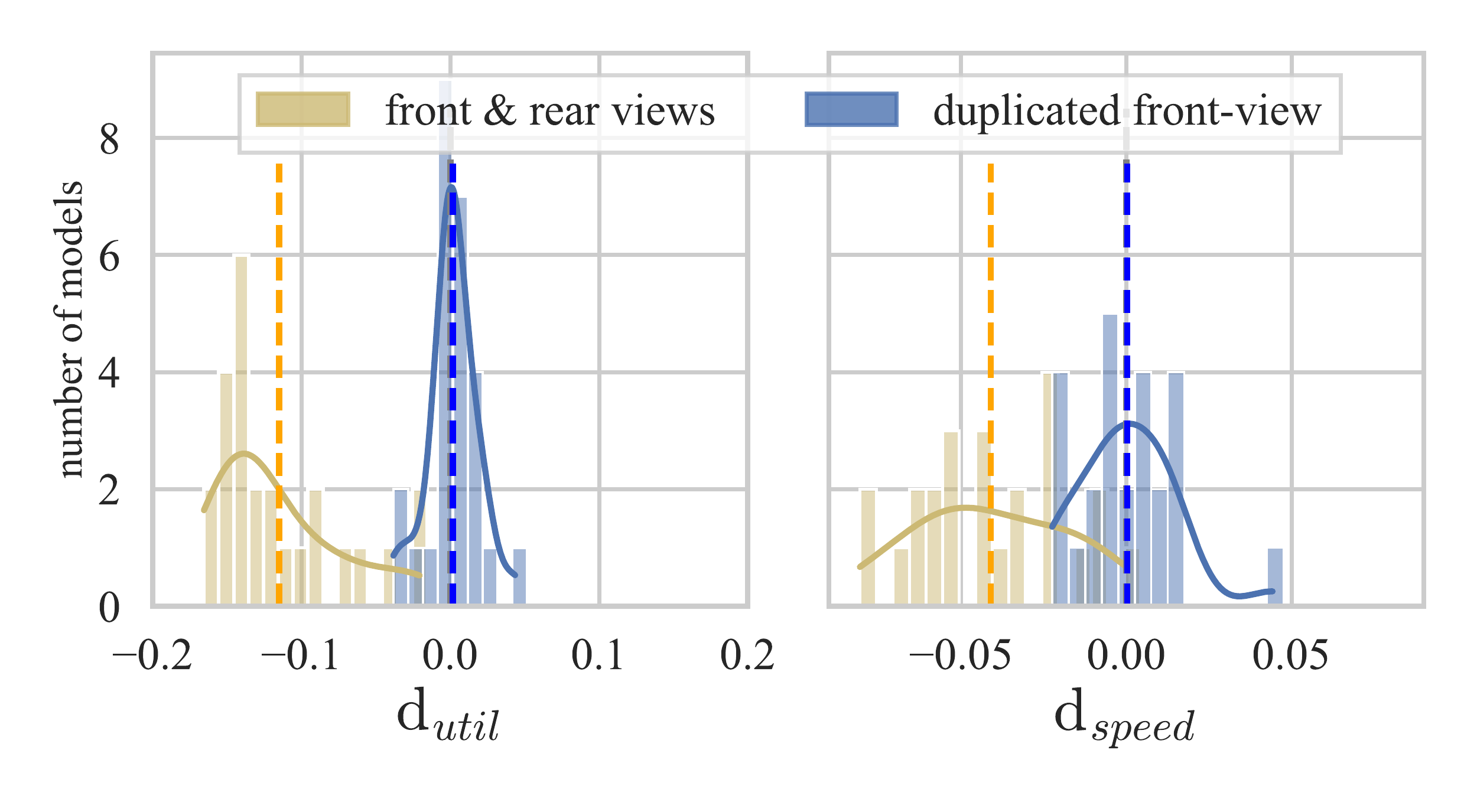}
         \caption{ModelNet40}
     \end{subfigure}
     \hspace*{\fill}
      \begin{subfigure}[b]{0.48\textwidth}
         \centering
         \includegraphics[width=1\textwidth, trim={10mm 10mm 5mm 5mm},clip]{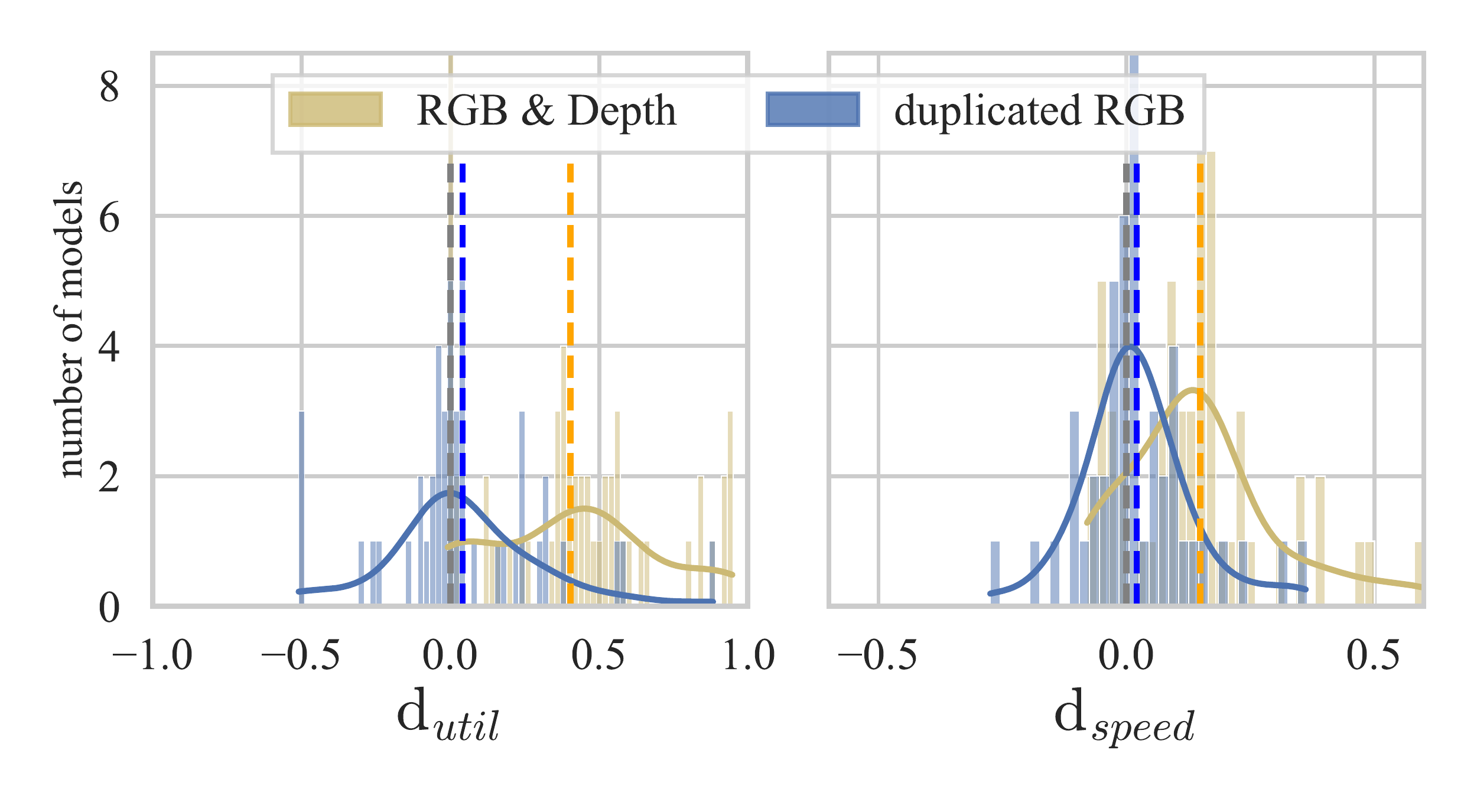}
         \caption{NVGesture}
     \end{subfigure}
\caption{Histograms and estimated density functions of $\widehat{\Diff}_{\util}$ and $\widehat{\Diff}_{\speed}$ of models trained using different modalities or duplicated single modality in (a) ModelNet40 or (b) NVGesture. We mark zero, $\mathbb{E}[\widehat{\Diff}_{\util}]$ and $\mathbb{E}[\widehat{\Diff}_{\speed}]$ with dashed lines. Many models have high $|\Diff_{\util}|$. We see $\widehat{\Diff}_{\util}$ is distributed asymmetrically around zero when using two different input modalities. When using two identical input modalities, it is distributed symmetrically around zero. We see $\widehat{\Diff}_{\speed}$ exhibits similar distributions as $\widehat{\Diff}_{\util}$.}
\label{fig:data_dup_nv}
\end{figure*}

\section{Experiments and Results}
\subsection{Datasets, Tasks and Baselines}
\label{sec:exp}

\textbf{Colored-and-gray-MNIST}~\citep{kim2019learning} is a synthetic dataset based on MNIST~\citep{lecun1998gradient}. In the training set of 60,000 examples, each example has two images, a gray-scale image and a monochromatic image, with color strongly correlated with its digit label. In the validation set of 10,000 examples, each example also has a gray-scale image and a corresponding monochromatic image, although with a low correlation between the color and its label. We consider the monochromatic image as the first modality $m_0$ and the gray-scale one as the second modality $m_1$. We use a neural network with four convolutional layers as the uni-modal branch and employ three MMTMs to connect them. The corresponding uni-modal DNNs trained on the monochromatic images and the gray-scale images achieve respective validation accuracies of 41\% and 99\%. We use this synthetic dataset mainly to demonstrate the proposed greedy learner hypothesis.

\textbf{ModelNet40} is one of the Princeton ModelNet datasets~\citep{wu20153d} with 3D objects of 40 categories (9,483 training samples and 2,468 test samples). The task is to classify a 3D object based on the 2D views of its front and back~\citep{su15mvcnn}.  
Each example is a collection of 2D images (224$\times$224 pixels) of a 3D object. For the uni-modal branches, we use ResNet18~\citep{he2016deep} and apply MMTMs in the three final residual blocks. The uni-modal DNNs achieve 89\% and 88\% accuracy when learning from the front view ($m_0$) and the rear view ($m_1$), respectively.

\textbf{NVGesture} (or NVIDIA Dynamic Hand Gesture Dataset~\citep{molchanov2015hand}) consists of 1,532 video clips (1,050 training and 482 test ones) of hand gestures divided into 25 classes. We sample 20\% of training examples as the validation set and use depth and RGB as the two modalities. We adopt the data preparation steps used in \citet{joze2020mmtm} and use the I3D architecture~\citep{carreira2017quo} as uni-modal branches and MMTMs as fusion modules in the six final inception modules. During training, we perform spatial augmentation on the video, including flipping and random cropping. During inference on the validation or test set, we perform center cropping on the video. 

We provide examples of each dataset and details on data preprocessing in \cref{app:data}.

\begin{figure*}[!b]
\begin{minipage}[t]{0.4\textwidth}
\centering
\includegraphics[width=0.8\linewidth,trim={10mm 5mm 5mm 5mm},clip]{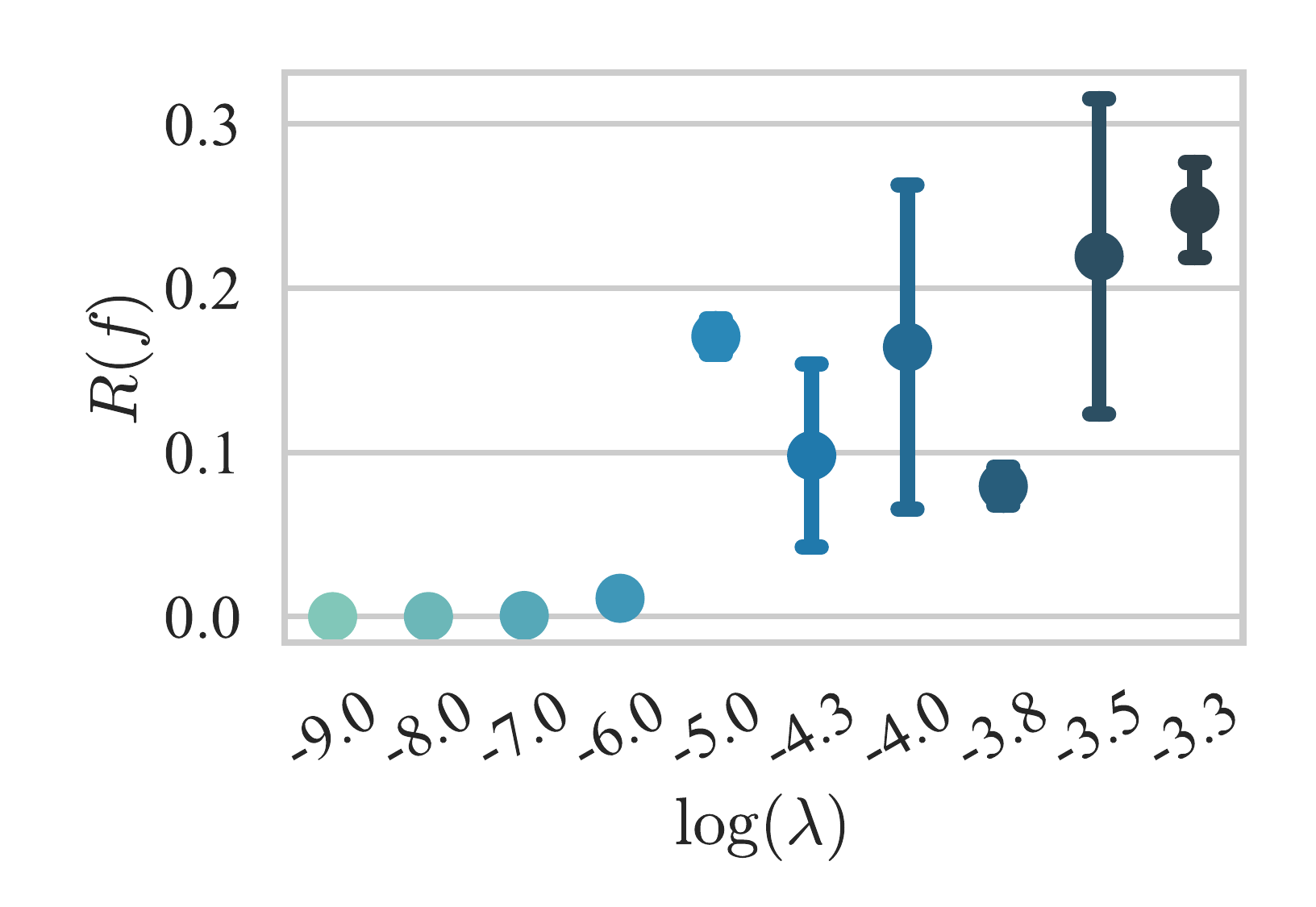}
\caption{The mean and standard deviation of $R(f)$ for models trained with different
weights ($\lambda$) on the L1 regularizer.. The larger the $\lambda$ is, the higher the $R(f)$ is, i.e., the sparser the model's parameters are.}
\label{fig:lambda_rf}
\end{minipage}
\hspace*{\fill} 
\begin{minipage}[t]{0.57\textwidth}
\centering
     \begin{subfigure}[b]{0.48\textwidth}
         \centering
\includegraphics[width=1\textwidth,trim={12mm 5mm 5mm 5mm},clip]{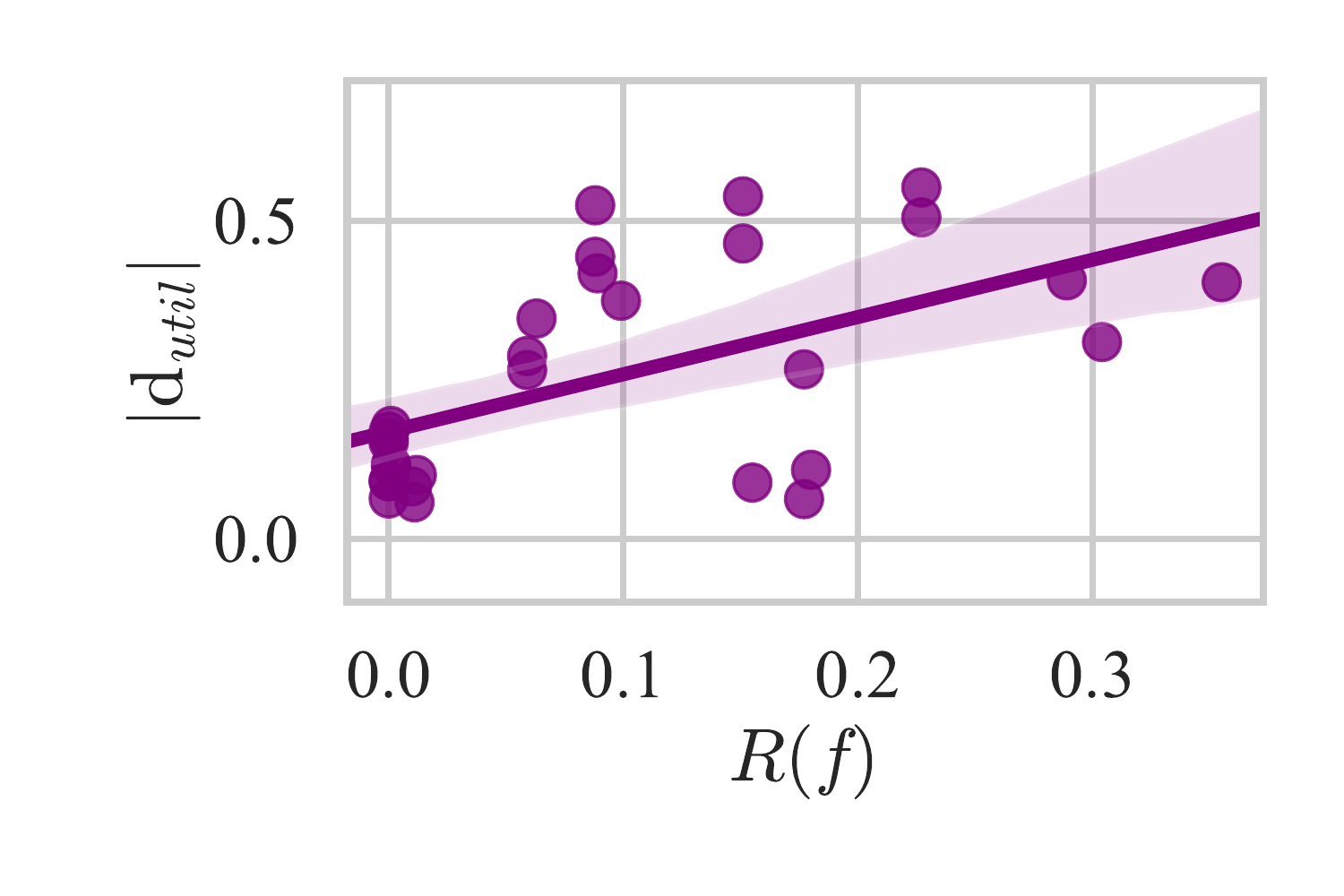}
\caption{$|\Diff_{\util}|$ vs. $R(f)$}
\end{subfigure}
\begin{subfigure}[b]{0.48\textwidth}
         \centering
\includegraphics[width=1\textwidth,trim={12mm 5mm 5mm 5mm},clip]{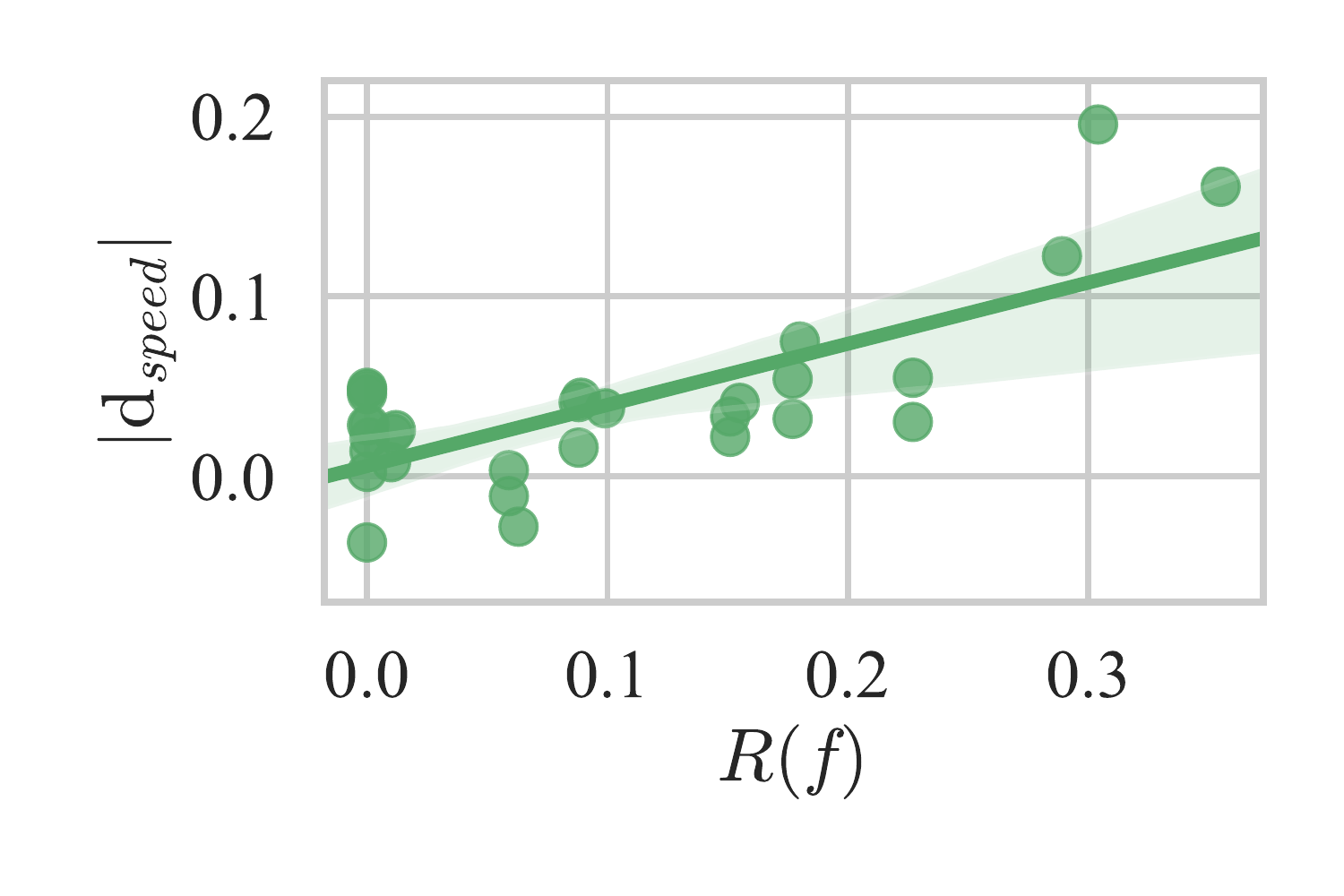}
\caption{$|\Diff_{\speed}|$ vs. $R(f)$}
\end{subfigure}
\caption{The observed $|\Diff_{\util}|$ and $|\Diff_{\speed}|$ for models trained with different weights ($\lambda$) on the L1 regularizer.
In (a) and (b), we see both $|\Diff_{\util}|$ and $|\Diff_{\speed}|$ increases along $R(f)$. It indicates that the multi-modal learning process becomes greedier if we introduce stronger regularization.} 
\label{fig:reg}
\end{minipage}
\end{figure*}

\subsection{Validating the Greedy Learner Hypothesis}
\label{sec:diff}

In this section, we run the conventional multi-modal learning process on tasks with different input and output pairs to illustrate our hypothesis experimentally. 
%We show that the multi-modal learning process cannot avoid producing models that exhibit a high degree of imbalance in utilization between modalities.  

For each task introduced in \S\ref{sec:exp}, in addition to the original dataset, we construct a dataset with two identical input modalities by duplicating one of the modalities. For example, when using the colored-and-gray-MNIST dataset, we predict the digit class using two identical gray-scale images. We train multi-modal DNNs on these datasets as explained below for each task: 
\begin{itemize}[leftmargin=*, topsep=-1pt, partopsep=-2pt]
    \item Colored-and-gray-MNIST: we train multi-modal DNNs using SGD with a momentum coefficient of 0.9 and a batch size of 128. We sample 20 learning rates at random from the interval $[10^{-5}, 1]$ on a logarithmic scale. We train the model four times using each of the learning rates and random initialization of the parameters. In total, we train 80 models. 
    \item ModelNet40: we use SGD without momentum and use a batch size of eight. We select nine learning rates from $10^{-3}$ to $1$ and train three copies of the model for each learning rate. This ends up with 27 models.
    \item NVGesture: we use a batch size of four, SGD with a momentum of 0.9, and uniformly sample 20 learning rates from the interval $[10^{-4}$, $10^{-1.5}]$ on a logarithmic scale. We train the model three times using each learning rate, resulting in 60 models in total. 
\end{itemize}

We present the results of the experiments on ModelNet40 and NVGesture in \cref{fig:data_dup_nv} and on Colored-and-gray-MNIST in \cref{fig:results_color} in \cref{app:figures}. We discuss three most interesting phenomena that we observe in the results. 

First, many models have high $|\Diff_{\util}|$. This confirms that the multi-modal learning process encourages the model to rely on one modality and ignore the other one, which is consistent with our hypothesis. We make this observation across all tasks, confirming that the conventional multi-modal learning process is greedy regardless of network architectures and tasks. 

Second, $\widehat{\Diff}_{\util}$ is distributed symmetrically around zero, and $\mathbb{E}[\widehat{\Diff}_{\util}]$ is approximately $0.0$, for all the experiments using two identical input modalities. On the other hand, if we use two distinct modalities, $\widehat{\Diff}_{\util}$ is distributed asymmetrically, and we observe $|\mathbb{E}[\widehat{\Diff}_{\util}]|$ of approximately 0.3, 0.1 and 0.4 for colored-and-gray-MNIST, ModelNet40 and NVGesture, respectively. We formalize the above observations as the following conjecture.

\emph{There exists an $\epsilon>0$, s.t.  $|\mathbb{E}[\widehat{\Diff}_{\util}]|>\epsilon$, when modalities are different. Otherwise, $\widehat{\Diff}_{\util}$ is distributed symmetrically around zero and $\mathbb{E}[\widehat{\Diff}_{\util}]=0$.}

Third, by observing conditional learning speed $\widehat{\Diff}_{\speed}$, we can draw the same conclusions. In fact, the distributions of $\widehat{\Diff}_{\speed}$ largely replicate the distributions of $\widehat{\Diff}_{\util}$. It validates our greedy learner hypothesis which attributes the imbalance in reliance on different modalities to the varying rate at which the learner learns from them. It moreover confirms $\Diff_{\speed}$ is an appropriate proxy to use to re-balance multi-modal learning.

\begin{table*}
\centering
\vspace{-1mm}
\caption{Test accuracy of best uni-modal DNNs and multi-modal DNNs trained using different strategies. 
}
\label{tab:guided-random}
\resizebox{0.94\linewidth}{!}{
\begin{threeparttable}
\begin{tabular}{lcccc}
  \toprule
   & Colored-and-gray-MNIST & ModelNet40 & NVGesture-scratch & NVGesture-pretrained 
    \\ \midrule
     uni-modal (best) &  99.14$\pm$0.11\tnote{1} & 89.34$\pm$0.39 & 77.59$\pm$0.55 & 78.98$\pm$2.02 \\ \midrule
     multi-modal (vanilla) & 45.26$\pm$0.46& 90.09$\pm$0.58  &  79.81$\pm$1.14 &  83.20$\pm$0.21 \\
       \, + RUBi~\citep{cadene2019rubi} & 44.79$\pm$0.62  &  90.45$\pm$0.58& 79.95$\pm$0.12  & 81.60$\pm$1.28\\
      \, + random (proposed)&  74.07$\pm$2.75  &  91.36$\pm$0.10 & 79.88$\pm$0.90 &  82.64$\pm$0.84 \\
      \, + guided (proposed) &  \textbf{91.01$\pm$1.20}  &  \textbf{91.37$\pm$0.28}  &  \textbf{80.22$\pm$0.73} & \textbf{83.82$\pm$1.45} \\
 \bottomrule
\end{tabular}
\begin{tablenotes}
\item[1]The monochromatic image cannot help with the prediction and it is hard to avoid it hurting the ensemble performance. The uni-modal branch taking gray-scale image in the multi-modal DNNs (guided) achieves an accuracy of 99.16$\pm$0.14. 
\vspace{-2mm}
\end{tablenotes}

\end{threeparttable}
}
\end{table*}
\subsection{Strong Regularization Encourages Greediness}
\label{sec:reg}
We also conjecture that regularization is a factor impacting the greediness of learning in multi-modal DNNs and strong regularization encourages greediness. We provide the following study to validate this. 

We investigate L1 regularization's impact on multi-modal DNNs. Precisely, we train the networks to optimize the loss $L' = L + \lambda ||\bm{\theta}||_1$, where $L$ is the classification loss in \S\ref{sec:formu}, $\bm{\theta}$ stands for all model parameters and $\lambda$ is the weight on the regularizer. 

We measure the effect of $\lambda$ on the network $f$ by computing the fraction of its parameters smaller than $10^{-7}$. We denote this quantity by $R(f)$. Since L1 regularization encourages sparsity of the network's parameters, as shown in \cref{fig:lambda_rf}, the larger the $\lambda$, the higher the $R(f)$. %We show that the stronger the regularization we apply to the DNNs' parameters during training, the higher the $|\mathbb{E}[\widehat{\Diff}_{\util}]|$ is. 

We conduct this study with ModelNet40, using the front and the rear views. We compute $|\Diff_{\util}(f)|$ for ten values of $\lambda$ from an interval $[10^{-9}, 10^{-3}]$. We use SGD without momentum, we set the learning rate to 0.1 and batch size to eight. Using each combination of hyperparameters, we repeat training for three times with random initialization and get three models. 

As shown in \cref{fig:reg}, $|\Diff_{\util}(f)|$ is positively correlated with $R(f)$, and when $\lambda \geq 10^{-5}$, $|\Diff_{\util}(f)|$ increases significantly with $\lambda$ increasing, as shown in \cref{fig:lambda_diff_both} in \cref{app:figures}. In other words, the stronger the regularization is, the larger the imbalance in utilization between modalities we observe. We also find that $|\Diff_{\speed}|$ follows the same trend as $|\Diff_{\util}|$. Again, it supports our choice of using the conditional learning speed to predict the conditional utilization rate.

\subsection{Balanced Multi-modal Learning}

Besides the proposed training algorithm (cf. \S\ref{sec:guided_algorithm}, referred to as \emph{guided}), we introduce a variant of it, referred to as \emph{random}: we perform regular steps in the first epoch, and afterwards, we let the model take a step that is randomly sampled from: regular step, re-balancing step on $m_0$, and re-balancing step on $m_1$. This algorithm is motivated by Modality Dropout~\citep{neverova2015moddrop} but better suited to multi-modal DNNs with intermediate fusion. We consider it a stronger baseline that can also balance learning from inputs of different modalities. 

\begin{figure}[h!]
\centering
     \begin{subfigure}[b]{0.235\textwidth}
         \centering
         \includegraphics[width=0.952\textwidth, trim={0mm 0mm 0mm 0},clip ]{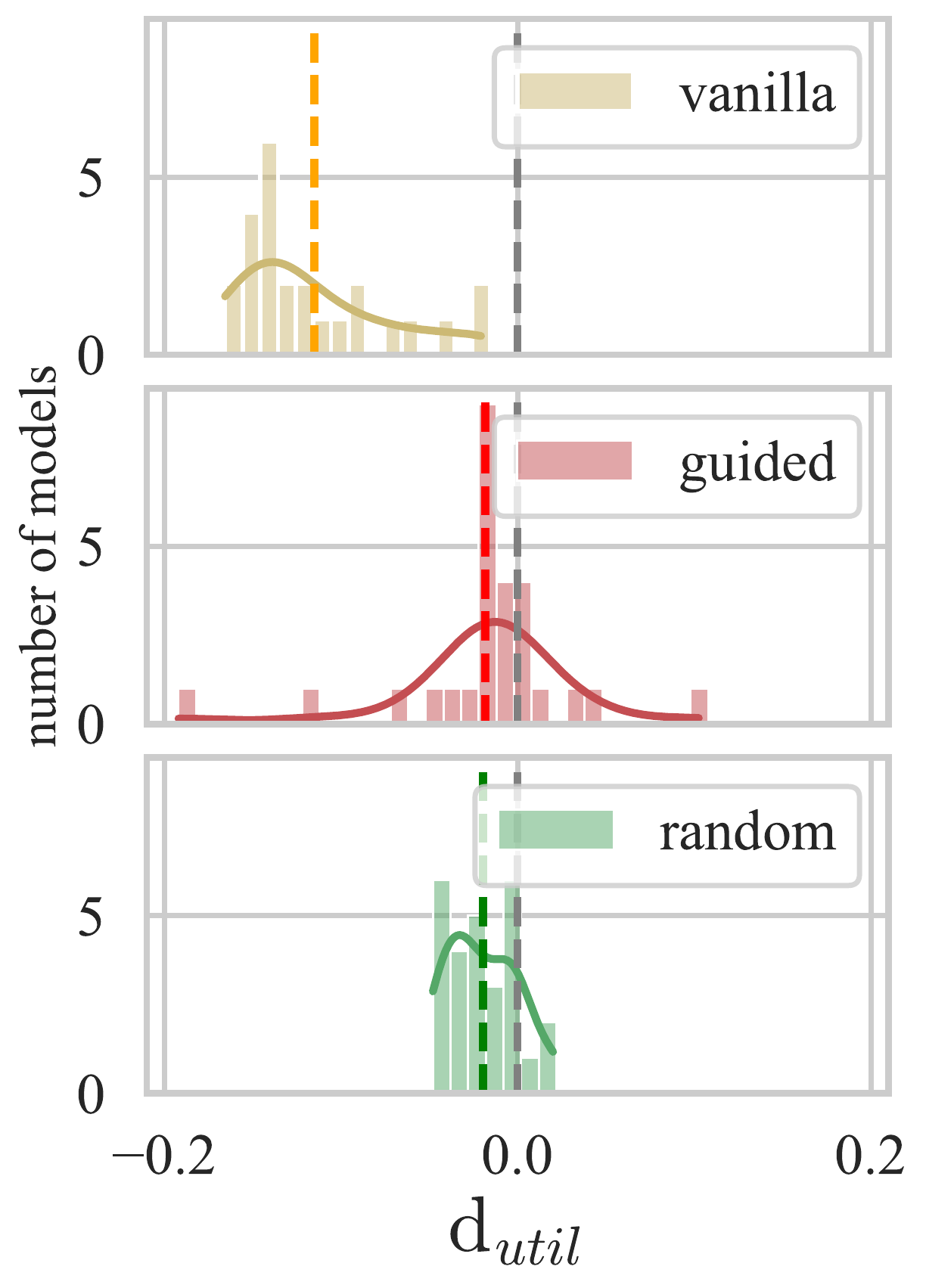}
         \caption{ModelNet40}
         \label{fig:perform-3d-1}
     \end{subfigure}
      \hspace*{\fill}
     \begin{subfigure}[b]{0.235\textwidth}
         \centering
         \includegraphics[width=1\textwidth, trim={0mm 0mm 0mm 0},clip]{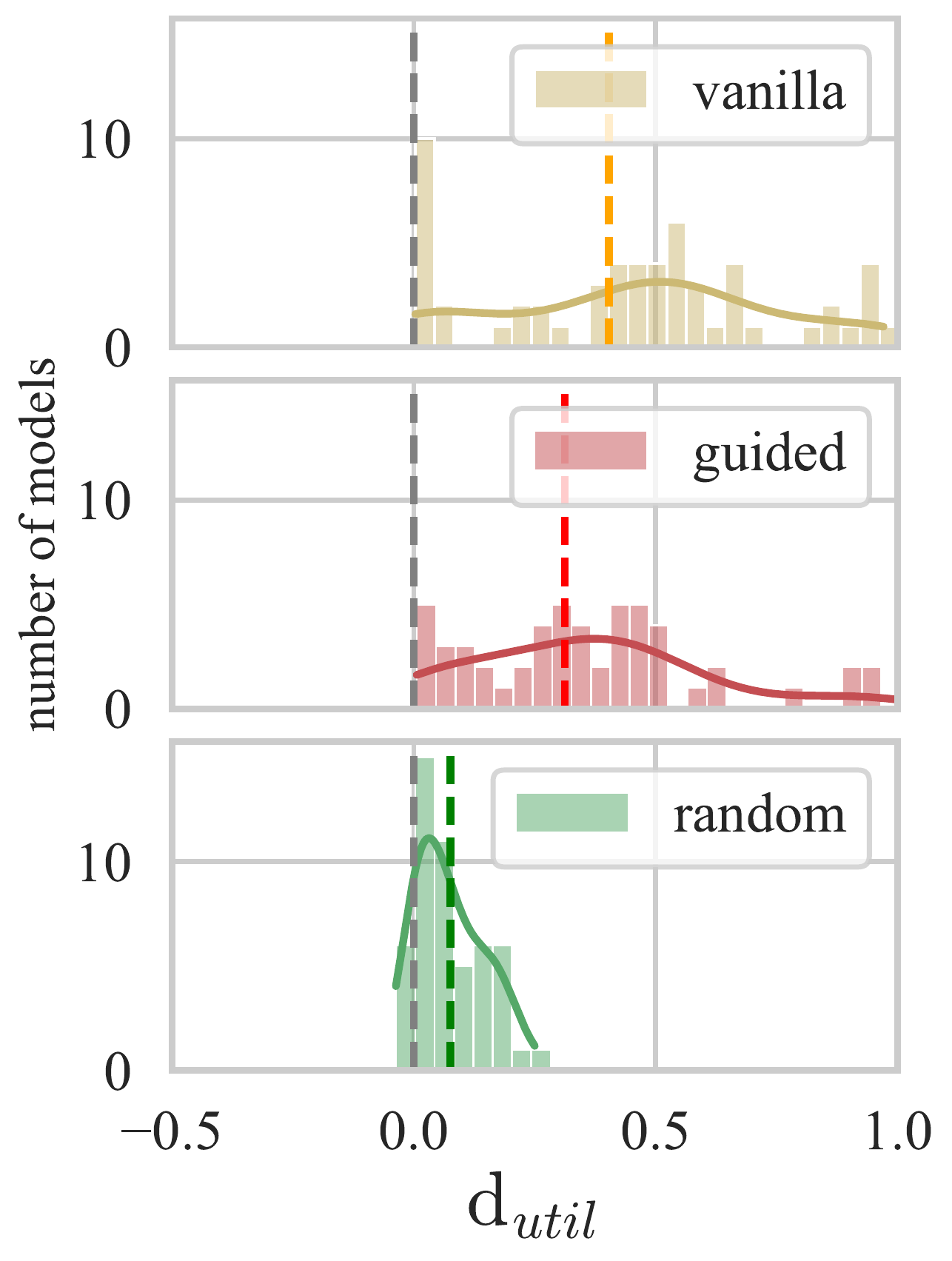}
         \caption{NVGesture}
         \label{fig:perform-nvgesture-1}
     \end{subfigure}
\caption{Histograms and estimated density functions of $\widehat{\Diff}_{\util}$ of models trained using the vanilla, the guided and the random algorithm on (a) ModelNet40 and (b) NVGesture. We use dashed lines to mark zero and $\mathbb{E}[\widehat{\Diff}_{\util}]$ over the set of models trained with each of the algorithms. Both versions of the proposed algorithm is less greedy than the vanilla one.}
\label{fig:performance-cal}
\end{figure}

\paragraph{Calibrated modality utilization}
We train multi-modal DNNs as described in \S\ref{sec:diff}, using the guided, the random, and the conventional training algorithm (referred to as \emph{vanilla}). For ModelNet40, we set the imbalance tolerance parameter $\alpha$ to 0.01 and the re-balancing window size $Q$ to 5. For NVGesture, we use $\alpha$ of 0.1 and $Q$ of 5.\footnote{We provide studies on the model's sensitivity to $Q$ and $\alpha$ in Figure~\ref{fig:hyper-Q} and Figure~\ref{fig:hyper-alpha} in \cref{app:figures}.} Results are shown in Figure~\ref{fig:performance-cal}. Models trained with the guided algorithm have lower $|\widehat{\Diff}_{\util}|$ compared to the vanilla algorithm. On NVGesture, we obtain $\mathbb{E}[\widehat{\Diff}_{\util}]$ of approximately 0.3 and 0.4 for models trained with the guided and the vanilla algorithm. On ModelNet40, we obtain $\mathbb{E}[\widehat{\Diff}_{\util}]$ of approximately -0.0, and -0.1 for models trained with the guided and the vanilla algorithm. 

The random version of the proposed method also calibrates modality utilization effectively, giving $\mathbb{E}[\widehat{\Diff}_{\util}]$ of approximately 0.1 and -0.0 for models trained on NVGesture and ModelNet40. We will see in the next section that it helps less on generalization compared to the guided version. 

\paragraph{Improved generalization performance}
We compare the generalization ability of multi-modal DNNs trained by the baseline (vanila), the proposed methods (guided and random) and the RUBi learning strategy~\citep{cadene2019rubi}. We chose RUBi as one of the baselines since it is model-agnostic and fusion-method-agnostic, and it is easy to adapt to different tasks. RUBi is designed to serve a similar purpose, that is, to encourage the model to learn from all modalities and has been shown to be helpful for VQA. 

For each algorithm, we train each model three times with the same learning rate. We use 0.01, 0.1 and 0.01 as learning rate for Colored-and-gray-MNIST, ModelNet40 and NVGesture respectively. We use $\alpha$ of 0.1 for Colored-and-gray-MNIST and NVGesture and 0.01 for ModelNet40. We set $Q$ of 5 for all three datasets. For NVGesture, we add one experiment where we initialize the model with parameters pre-trained using the Kinetics dataset~\citep{carreira2017quo} in addition to random initialization. We refer to this setting as ``NVGesture-pretrained'' and to the other one as ``NVGesture-scratch''. 

Besides the multi-modal DNNs trained using different strategies, we implement Squeeze-and-Excitation blocks~\citep{hu2018squeeze} to obtain uni-modal DNNs that are comparable to the multi-modal DNNs with MMTMs~\citep{joze2020mmtm}. We train uni-modal DNNs with each modality and present the best performance they achieve for each task. We report means and standard deviations of the models' test accuracies in \cref{tab:guided-random}. The guided algorithm improves the models' generalization performance over all other three methods in all four cases.\footnote{Results on NVGesture are not directly comparable to numbers in other works since we use 20\% training samples as the validation data.} RUBi does not show consistent improvement across tasks compared to the vanilla algorithm. We did not find this strategy as helpful as reported for VQA. 

Our idea can be extended naturally to tasks with more than two modalities. The conditional learning speed of the $i^{th}$ modality can be derived using the weight norm and gradient norm of the $i^{th}$ uni-modal network's parameters and the fusion module's parameters impacting $\hat{y}_i$. The conditional utilization rate would be computed analogously to the bi-modal case, by passing the averaged feature maps of all other modalities to the fusion modules. We define $\Diff_{\speed}$ by the two most different conditional learning speeds and accelerate the learning of one modality per time. We experiment with three views in ModelNet40 and find that the guided algorithm outperforms the vanilla one as it improves the accuracy from  $91.21\%$ to $92.45\%$.

In summary, we present that the greediness of the current training algorithm is an issue preventing multi-modal DNNs from achieving better performance. The proposed method can help to overcome this issue on multiple datasets. 

\section{Discussion}
Our work demonstrates that the end-to-end trained multi-modal DNNs rely on one of the input modalities to make predictions while leaving the other modalities underutilized. For many multi-modal classification tasks, such as video classification, the harm of this behavior may not be as obvious as in VQA which strongly relies on cross-modal reasoning. However, the tools we propose in this work enable us to have a concrete look of what the model has learned, which has been a missing component in multi-modal learning systems. We validated our statements experimentally on three multi-modal datasets and illustrated that using the proposed algorithm to balance the models' learning from different modalities enhances generalization. This result emphasizes that the adequate modality utilization is a desired property that a model should achieve in multi-modal learning. In addition to the previous discussion on this problem, our greedy learner hypothesis provides a complementary explanation and the methods inspired by it enrich the spectrum of available tools for multi-modal learning. 
\section*{Acknowledgements}
This work was supported in part by grants from the National Institutes of Health (P41EB017183 and R21CA225175), the National Science Foundation (HDR-1922658), the Gordon and Betty Moore Foundation (9683), and Samsung Advanced Institute of Technology (under the project \textit{Next Generation Deep Learning: From Pattern Recognition to AI}). Nan Wu is supported by Google PhD Fellowship.
We thank Catriona Geras and Taro Makino for their comments on writing. We appreciate the suggestions from ICML reviewers.

\bibliography{example_paper}
\bibliographystyle{icml2022}

%%%%%%%%%%%%%%%%%%%%%%%%%%%%%%%%%%%%%%%%%%%%%%%%%%%%%%%%%%%%%%%%%%%%%%%%%%%%%%%
%%%%%%%%%%%%%%%%%%%%%%%%%%%%%%%%%%%%%%%%%%%%%%%%%%%%%%%%%%%%%%%%%%%%%%%%%%%%%%%
% APPENDIX
%%%%%%%%%%%%%%%%%%%%%%%%%%%%%%%%%%%%%%%%%%%%%%%%%%%%%%%%%%%%%%%%%%%%%%%%%%%%%%%
%%%%%%%%%%%%%%%%%%%%%%%%%%%%%%%%%%%%%%%%%%%%%%%%%%%%%%%%%%%%%%%%%%%%%%%%%%%%%%%
\newpage
\appendix
\onecolumn

\section{Data Preparation}
\label{app:data}
\begin{figure*}[htb!]
    \centering
    \includegraphics[width=0.76\textwidth]{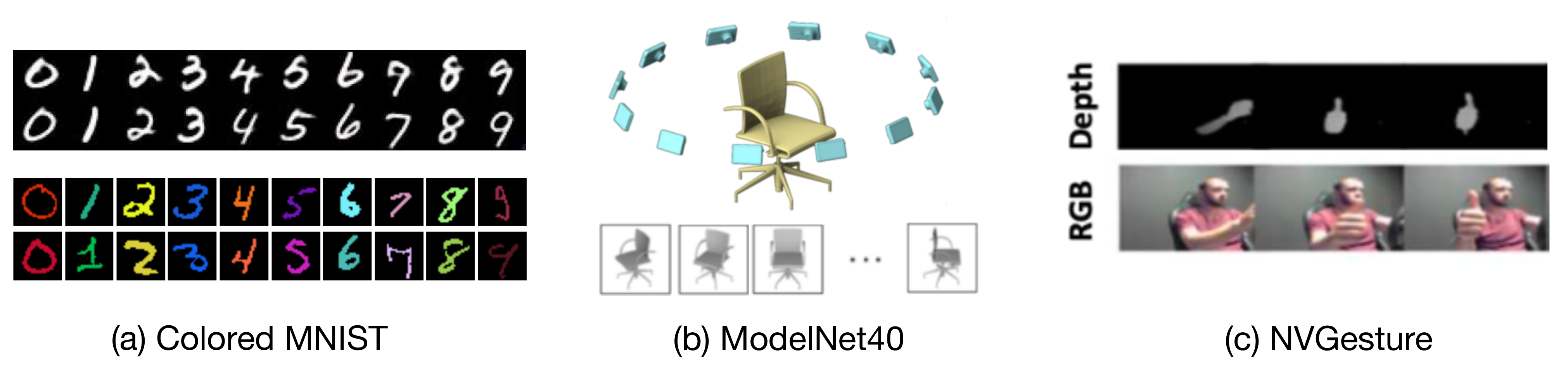}
    \caption{(a) The Colored MNIST dataset~\citep{kim2019learning}. We consider the monochromatic image, and the gray-scale image as the two input modalities (b) The ModelNet40 dataset~\citep{su15mvcnn}. The 2D representations are gray-scale images rendered from 12 different viewpoints of the object. 
    (c)The NVGesture dataset~\citep{molchanov2015hand}. We use depth and RGB channels as the two modalities.}
    \label{fig:multi-modal-data}
\end{figure*}

We used three datasets in the paper: Colored MNIST dataset~\citep{kim2019learning}, ModelNet40 dataset~\citep{su15mvcnn} and NVGesture dataset~\citep{molchanov2015hand}, as illustrated in \cref{fig:multi-modal-data}. 

For Colored MNIST and ModelNet40, we did not perform any extra data pre-processing steps on the original datasets. 

For the NVGesture dataset, each video has a resolution of 240$\times$320 and a duration of 80 frames from action starting to ending. There are three videos with unmatched starting indices between RGB and depth. We adopt the starting frame indice of RGB for all modalities. We randomly select 64 consecutive frames from the videos in the dataset and if the video has less than 64 frames, we zero-pad on both sides of it to obtain 64 frames. Frames are resized as 256$\times$256 and are cropped into 224$\times$224 as inputs (we use static cropping where we crop from the same location across times and modalities).

\newpage
\section{Supplementary Figures}
\label{app:figures}
\begin{figure*}[hbt!]
\centering
\includegraphics[width=0.8\linewidth]{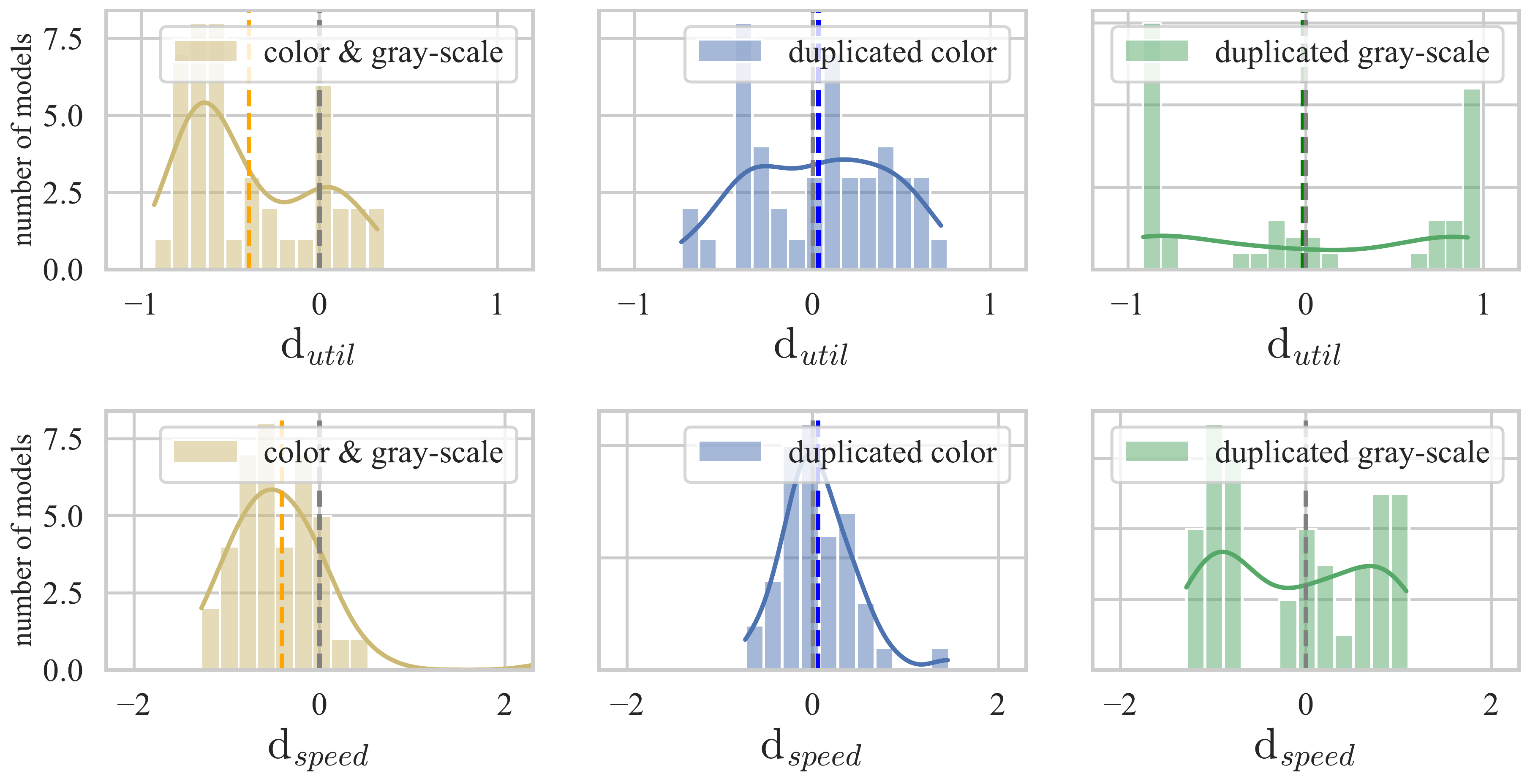}
\caption{Histograms and estimated density functions of $\Diff_{\util}$ and $\Diff_{\speed}$ of models trained for colored-and-gray-MNIST, using monochromatic and gray-scale images as two modalities, using identical monochromatic images as two modalities and using identical gray-scale images as two modalities. Note that $\Diff_{\speed}$ is not bounded by 1 as $\Diff_{\util}$ is. Thus when learning from the two modalities is happening at very different paces, we can observe patterns shown in this study.}
\label{fig:results_color}
\end{figure*}

\begin{figure}[hbt!]
\centering
\includegraphics[width=0.43\linewidth,trim={10mm 5mm 0 0},clip]{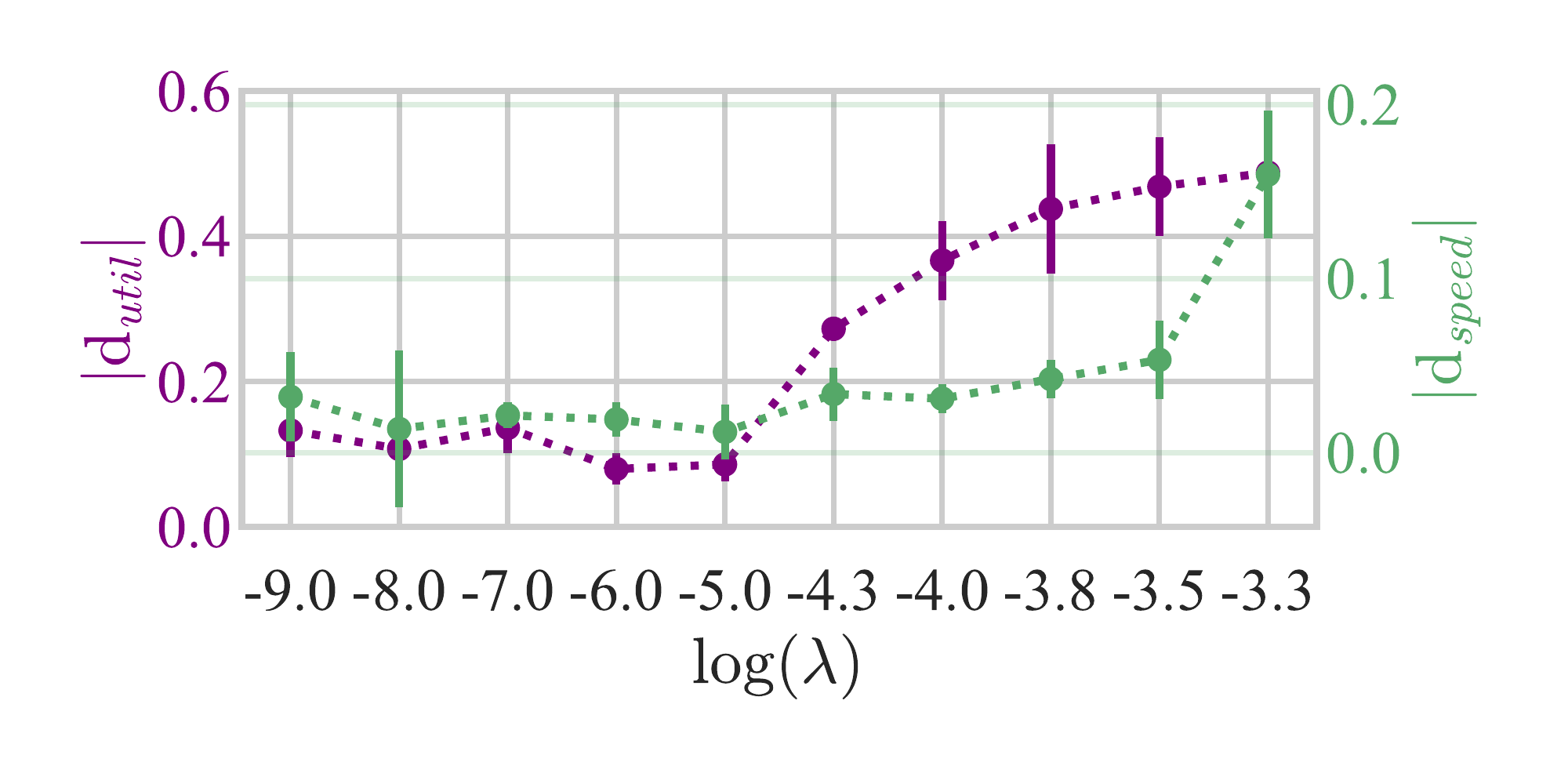}
\caption{The observed $|\Diff_{\util}|$ and $|\Diff_{\speed}|$ for models trained with different weights ($\lambda$) on the L1 regularizer. We show $|\Diff_{\util}|$ and $|\Diff_{\speed}|$ as a function of $\log(\lambda)$. As we can see, when $\log(\lambda)\geq-5$, both $|\Diff_{\util}(f)|$ and $|\Diff_{\speed}|$ increases with $\lambda$ increasing, .}
\label{fig:lambda_diff_both}
\end{figure}

\begin{figure*}[hbt!]
\centering
     \begin{subfigure}[t]{0.48\textwidth}
         \centering
     \includegraphics[width=1\textwidth, trim={10mm, 5mm, 0, 0}, clip]{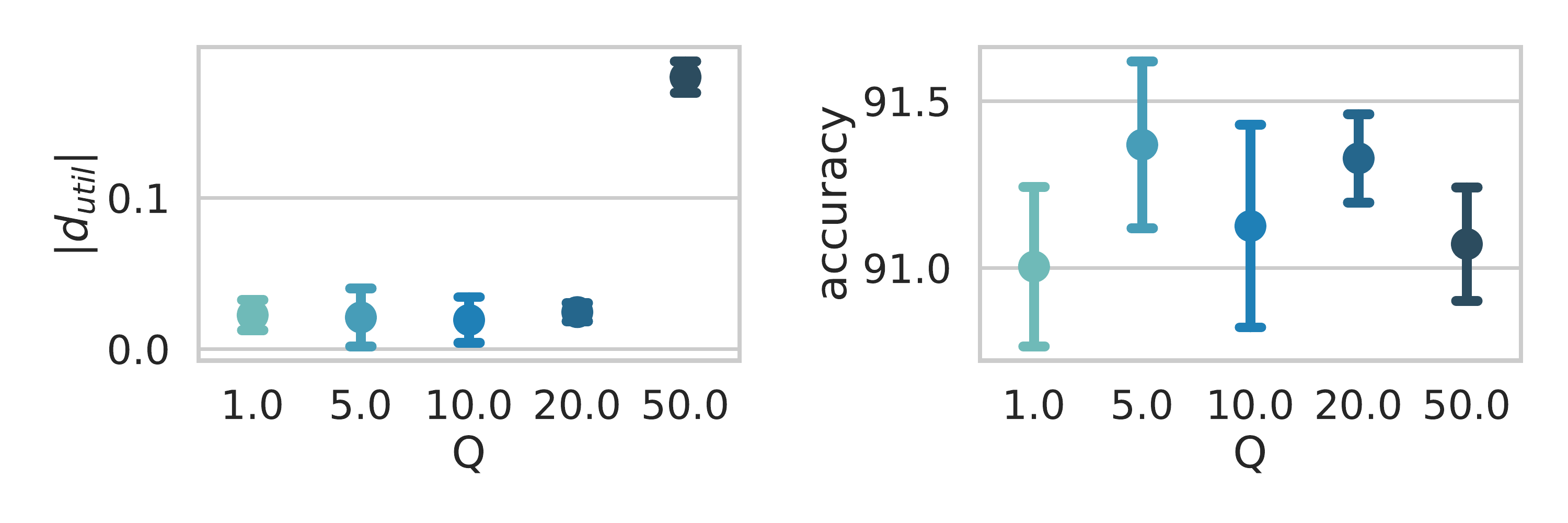}
        \caption{We train models using $Q \in \{1, 5, 10, 20, 50\}$ and $\alpha=0.01$. Except $Q=50$, models trained using other values of $Q$ present relatively balanced conditional utilization rates between modalities (left panel). According to the models' generalization performance (right panel), we choose $Q=5$ for the final experiment.}
        \label{fig:hyper-Q}
     \end{subfigure}
     \hspace*{\fill} 
     \begin{subfigure}[t]{0.48\textwidth}
         \centering
        \includegraphics[width=1\textwidth, trim={10mm, 5mm, 0, 0}, clip]{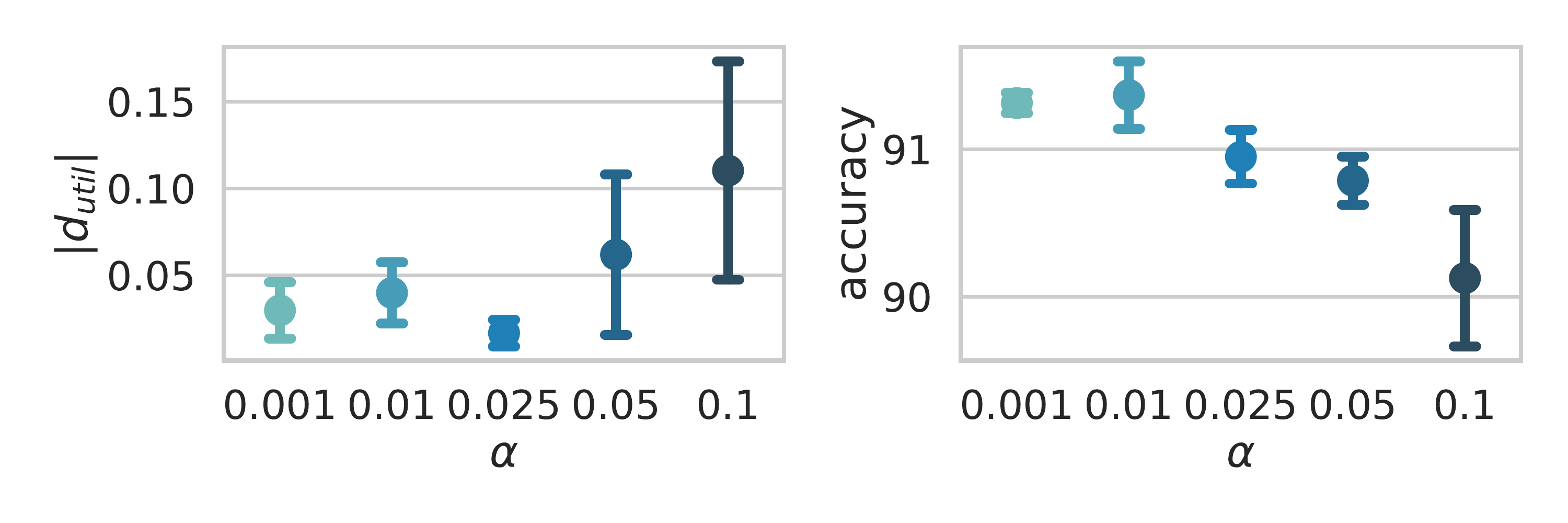}
        \caption{We train models using $Q=5$ and $\alpha \in  \{0.01, 0.1, 0.25, 0.5, 1\} \times \mathbb{E}[\widehat{\Diff}_{\util}]$, where $\mathbb{E}[\widehat{\Diff}_{\util}]=0.1$ as shown in \S\ref{sec:diff}.
        When setting $\alpha \leq 0.25\mathbb{E}[\widehat{\Diff}_{\util}]$, we can control $\Diff_{\util}$ effectively (left panel). We choose to use $\alpha=0.1\mathbb{E}[\widehat{\Diff}_{\util}]=0.01$ based on the models' accuracy (right panel) for the final experiment.}
        \label{fig:hyper-alpha}
        \end{subfigure}
    \caption{Models' behavior when using different values for the imbalance tolerance parameter $\alpha$ and the re-balancing window size $Q$ in the balanced multi-modal training algorithm. We use ModelNet40 (front and rear views) and fix the learning rate at 0.1 for the studies on both hyperparameters.}
\end{figure*}

\end{document}